\documentclass[final,3p,times,onecolumn]{elsarticle}
\usepackage{graphicx}
\usepackage[export]{adjustbox}
\usepackage{float}
\usepackage{paralist}
\usepackage{multirow}
\usepackage[ruled,vlined]{algorithm2e}
\usepackage{pifont}
\usepackage{textcomp}
\usepackage{comment}
\usepackage{amsmath}
\usepackage{amssymb}
\usepackage{url}

\usepackage{tabularx}
\newcolumntype{C}{>{\centering\arraybackslash}X}
\newcolumntype{R}{>{\raggedleft\arraybackslash}X}
\newcolumntype{L}{>{\raggedright\arraybackslash}X}

\makeatletter
\g@addto@macro{\UrlBreaks}{\UrlOrds} 
\makeatother
\usepackage[hang,flushmargin]{footmisc} 
\usepackage[utf8]{inputenc}
\AtBeginDocument{}
\usepackage{color}

\journal{Applied Soft Computing}\usepackage[T1]{fontenc}
\usepackage[utf8]{inputenc} 
\usepackage{comment}
\usepackage{booktabs}
\usepackage{xargs}
\usepackage[pdftex,dvipsnames]{xcolor}
\usepackage[colorinlistoftodos,prependcaption,textsize=tiny]{todonotes}
\usepackage{tikz}

\newcommandx{\unsure}[2][1=]{\todo[linecolor=red,backgroundcolor=red!25,bordercolor=red,#1]{#2}}
\newcommandx{\change}[2][1=]{\todo[linecolor=blue,backgroundcolor=blue!25,bordercolor=blue,#1]{#2}}
\newcommandx{\info}[2][1=]{\todo[linecolor=OliveGreen,backgroundcolor=OliveGreen!25,bordercolor=OliveGreen,#1]{#2}}
\newcommandx{\improvement}[2][1=]{\todo[linecolor=Plum,backgroundcolor=Plum!25,bordercolor=Plum,#1]{#2}}
\newcommandx{\thiswillnotshow}[2][1=]{\todo[disable,#1]{#2}}

\begin{document}
\begin{frontmatter}

\title{Advanced Machine Learning Techniques for Fake News (Online Disinformation) Detection: A Systematic Mapping Study}

\author[UTP]{Micha\l{} Chora\'s} \ead{chorasm@utp.edu.pl}
\author[NTU]{Konstantinos Demestichas} \ead{cdemest@cn.ntua.gr}
\author[UTP]{Agata Gie\l{}czyk}\ead{agata.gielczyk@utp.edu.pl}
\author[UBU]{\'Alvaro Herrero}\ead{ahcosio@ubu.es}
\author[PWR]{Pawe\l{} Ksieniewicz}\ead{pawel.ksieniewicz@pwr.edu.pl}
\author[NTU]{Konstantina Remoundou}\ead{kremoundou@cn.ntua.gr}
\author[UBU]{Daniel Urda}\ead{durda@ubu.es}
\author[PWR]{Micha\l{} Wo\'zniak\corref{cor1}}\ead{michal.wozniak@pwr.edu.pl}

\cortext[cor1]{Corresponding author}
\address[UTP]{UTP University of Science and Technology, Poland}
\address[NTU]{National Technical University of Athens, Greece}
\address[UBU]{Grupo de Inteligencia Computacional Aplicada (GICAP), 
Departamento de Ingenier{\'i}a Inform{\'a}tica, 
Escuela Polit{\'e}cnica Superior, 
Universidad de Burgos, Av. Cantabria s/n, 09006, 
Burgos, Spain.}
\address[PWR]{Wroc\l{}aw University of Science and Technology, Poland}


\begin{abstract}
Fake news has now grown into a big problem for societies and also a major challenge for people fighting disinformation. This phenomenon plagues democratic elections, reputations of individual persons or organizations, and has negatively impacted citizens, (e.g., during the COVID-19 pandemic in the US or Brazil). Hence, developing effective tools to fight this phenomenon by employing advanced Machine Learning (ML) methods poses a significant challenge. The following paper displays the present body of knowledge on the application of such intelligent tools in the fight against disinformation. It starts by showing the historical perspective and the current role of fake news in the information war. Proposed solutions based solely on the work of experts are analysed and the most important directions of the application of intelligent systems in the detection of misinformation sources are pointed out. Additionally, the paper presents some useful resources (mainly datasets useful when assessing ML solutions for fake news detection) and provides a short overview of the most important R\&D projects related to this subject. The main purpose of this work is to analyse the current state of knowledge in detecting fake news; on the one hand to show possible solutions, and on the other hand to identify the main challenges and methodological gaps to motivate future research.


\end{abstract}


\end{frontmatter}

\section{Introduction}
Let us start with a strong statement: the fake news phenomenon is currently a big problem for societies, nations and individual citizens. Fake news has already plagued democratic elections, reputations of individual persons or organizations, and has negatively impacted citizens in the COVID-19 pandemic (e.g., fake news on alleged medicines in the US or in Brazil). It is clear we need agile and reliable solutions to fight and counter the fake news problem. Therefore, this article demonstrates a critical scrutiny of the present level of knowledge in fake news detection, on one hand to show possible solutions but also to motivate the future research in this domain.

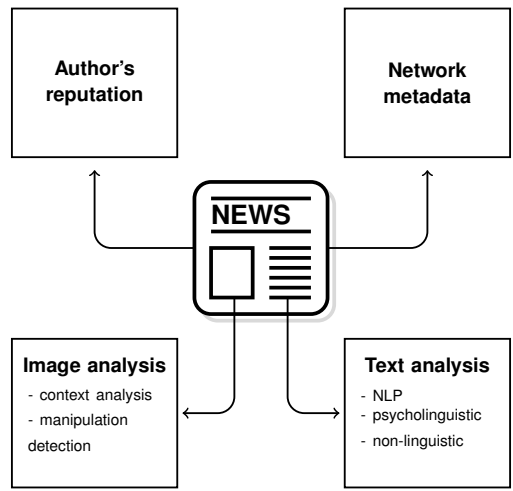
\begin{figure}
    \centering

\resizebox{.4\textwidth}{!}{%
\begin{tikzpicture}[font=\sffamily\small,
cir/.style={draw,text width=1.8cm,minimum height=2cm, rounded corners=1cm, align=center,inner sep=.1cm, draw=black, fill=white, text=black, thick},
ar/.style={shorten >= 2pt, thick, rounded corners=.25cm},
]


\draw[black!15, rounded corners=.25cm, ultra thick, fill=white] (-.9,-1.1) rectangle (1.1,.9);

\draw[black, rounded corners=.25cm, ultra thick, fill=white] (-1,-1) rectangle (1,1);
\node[align=left, text width=1.5cm] at (0, .5) {\large \bfseries NEWS};
\draw[-, ultra thick] (-.75,.75) -- (.75,.75);
\draw[-, ultra thick] (-.75,.25) -- (.75,.25);
\draw[ultra thick] (-.75,0) rectangle (-.125,-.75);

\draw[-, ultra thick] (.125,0) -- (.75,0);
\draw[-, ultra thick] (.125,-.15) -- (.75,-.15);
\draw[-, ultra thick] (.125,-.3) -- (.75,-.3);
\draw[-, ultra thick] (.125,-.45) -- (.75,-.45);
\draw[-, ultra thick] (.125,-.6) -- (.75,-.6);
\draw[-, ultra thick] (.125,-.75) -- (.75,-.75);

\node[draw, text width=2.25cm, minimum height=2.25cm, thick, align=center] at (-2.5,2.5) {\bfseries Author's\\ reputation};
\node[draw, text width=2.25cm, minimum height=2.25cm, thick, align=center] at (2.5,2.5) {\bfseries Network\\ metadata};

\node[draw, text width=2.25cm, minimum height=2.25cm, thick, align=center] at (2.5,-2.5) {};
\node[text width=2.1cm, minimum height=2cm, thick, align=center] at (2.5,-1.8) {\bfseries Text analysis};
\node[text width=2cm, minimum height=2cm, thick, align=left] at (2.5,-2.6) {\scriptsize - NLP\\- psycholinguistic\\- non-linguistic};

\node[draw, text width=2.25cm, minimum height=2.25cm, thick, align=center] at (-2.5,-2.5) {};
\node[text width=2.25cm, minimum height=2cm, thick, align=center] at (-2.5,-1.8) {\bfseries Image analysis};
\node[text width=2cm, minimum height=2cm, thick, align=left] at (-2.5,-2.6) {\scriptsize - context analysis\\- manipulation detection};

\draw[->, ar] (-1, 0) -- (-2.5,0) -- (-2.5,1.25);
\draw[->, ar] (1, 0) -- (2.5,0) -- (2.5,1.25);

\draw[->, ar] (-.4, -.75) -- (-.4,-2.5) -- (-1.25,-2.5);
\draw[->, ar] (.4, -.75) -- (.4,-2.5) -- (1.25,-2.5);





\end{tikzpicture}}
    \caption{The types of digital content that are analysed so as to detect fake news in an automatic manner}
    \label{fig:approaches}
\end{figure}

Fake news is a tough challenge to overcome, however there are some efforts from the Machine Learning (ML) community to stand up to this harmful phenomenon. In this mapping study, we present such efforts, solutions and ideas. As it is presented in Fig.~\ref{fig:approaches}, fake news detection may be performed by analysing several types of digital content such as images, text and network data, as well as the author/source reputation.

This survey is not the first one in the domain of fake news. Another major comprehensive work addressing the ways to approach fake news detection (mainly text analysis-based) and mainstream fake news datasets is~\cite{parikh2018media}. According to it, the \emph{state-of-the-art} approaches for this kind of analysis may be classified into five general groups with methods relying upon: (\emph{i}) linguistic features, (\emph{ii}) deception modelling, (\emph{iii}) clustering, (\emph{iv}) predictive modelling and (\emph{v}) content cues. With regard to the text characteristics, style-based and pattern-based detection methods are also presented in~\cite{Zhou:2018}. Those methods rely on the analysis of specific language attributes and the language structure. The analyzed attributes found by the authors of the survey include such features as: quantity of the language elements (e.g. verbs, nouns, sentences, paragraphs), statistical assessment of language complexity, uncertainty (e.g. number of quantifiers, generalizations, question marks in the text), subjectivity, non-immediacy (such as the count of rhetorical questions or passive voice), sentiment, diversity, informality and specificity of the analyzed text. Paper \cite{conroy2015automatic} surveys several approaches to assessing fake news, which stem from two primary groups: linguistic cue approaches (applying ML) as well as network analysis approaches. 



Yet another category of solutions is network-based analysis. In \cite{zubiaga2018detection}, two distinct categories are mentioned: (\emph{i}) social network behavior analysis to authenticate the news publisher's social media identity and to verify their trustworthiness and (\emph{ii}) scalable computational fact-checking methods based on knowledge networks. Beside text-based and network-based analysis, some other approaches are reviewed. For example, \cite{sharma2019combating} attempts to survey identification and mitigation techniques in combating fake news and discusses feedback-based identification approaches. 


Crowd-signal based methods are also reported in \cite{tschiatschek2018fake}, while content propagation modelling for fake news detection purposes, alongside credibility assessment methods, are discussed in \cite{Zhou:2018}. Such credibility-based approaches are categorized here into four groups: evaluation of news headlines, news source, news comments and news spreaders/re-publishers. In addition, in some surveys, content-based approaches using non-text analysis are discussed. The most common ones are based on image analysis \cite{parikh2018media, sharma2019combating}.

As complementary to the mentioned surveys, the present paper is unique by catching a very different angle of fake news detection methods (focused on advanced ML approaches). Moreover, in addition to overviewing current methods, we propose our own analysis criterion and categorization. We also suggest expanding the context of methods applicable for such a task and describe the datasets, initiatives and current projects, as well as the future challenges.

The remainder of the paper is structured in the following manner: in Section 1, previous surveys are overviewed and the historic evolution of fake news is presented, its current impact as well as the problem with definitions. In Section 2, we present current activities to address the fake news detection problem as well as technological and educational actions. Section 3 constitutes the in-depth systematic mapping of ML based fake news detection methods focused on the analysis of text, images, network data and reputation. 
Section 4 describes the relevant datasets used nowadays. In the final part of the paper we present some most emerging challenges in the discussed domain and we draw the main conclusions. 


\subsection{A historic perspective}

Even though the fake news problem has lately become increasingly important, it is not a recent phenomenon. According to different experts \cite{posetti2018short}, its origins are in ancient times. 
The oldest recorded case of spreading lies to gain some advantage is the disinformation campaign that took place on the eve of the \emph{Battle of Kadesh}, dated around 1280 B.C., where the Hittite Bedouins deliberately got arrested by the Egyptians in order to tell \emph{Pharaoh Ramses II} the wrong location of the \emph{Muwatallis II} army \cite{cline20151177}. 

Long time after that, in 1493, the Gutenberg printing press was invented. This event is widely acknowledged as a keystone in the history of news and press media, as it revolutionized this field. As a side effect, the dis- and misinformation campaigns had immeasurably intensified. As an example, it is worth mentioning the \emph{Great Moon Hoax}, dating back to 1835. This term is a reference to the collection of half a dozen papers published in \emph{The Sun}, the newspaper from New York. These articles concerned the ways life and culture had allegedly been found on the Moon. 

More recently, fake news and disinformation played a crucial role in World War I and II. On the one hand, British propaganda during World War I was aimed at demonising German enemies, accusing them of using the remains of their troops to obtain bone meal and fats, and then feeding the rest to swines. As a negative consequence of that, Nazi atrocities during World War II were initially doubted \cite{neander2010media}.

On the other hand, fake news was also generated by Nazis for sharing propaganda. Joseph Goebbels, who cooperated closely with Hitler and was responsible for German Reich's propaganda, performed a deciding role in the news media of Germany. He ordered the publication of a paper \emph{The Attack}, which was then used to disseminate brainwashing information. By means of untruth and misinformation, the opinion of the public was being persuaded to be in favour of the dreadful actions of the Nazis. Furthermore, according to \cite{herzstein1978most}, to this day, it has been the most disreputable propaganda campaign ever mounted.

\begin{figure}[!ht]
    \centering
    \includegraphics[width=.8\textwidth]{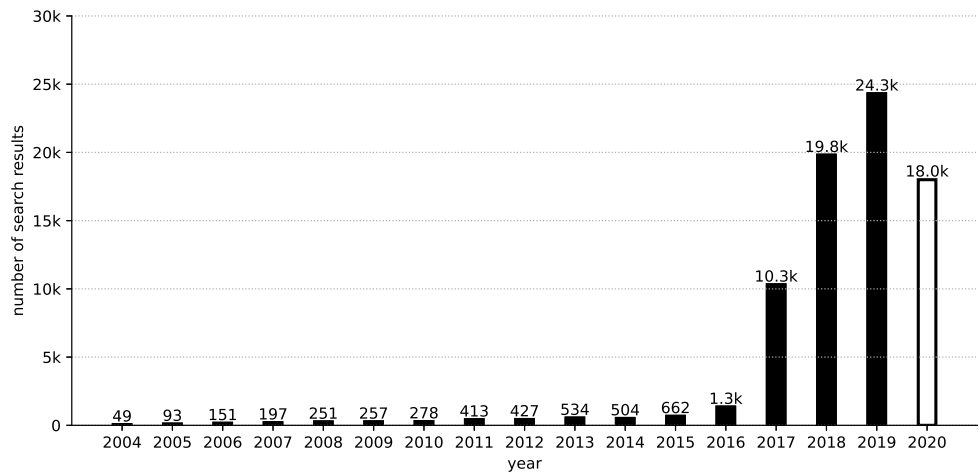}
    \caption{Evolution of the number of publications \emph{per year} retrieved from the keyword "\emph{fake news}" according to \emph{Google Scholar}. For the year 2020, the status as of September, 8th.}
    \label{fig:scholar}
\end{figure}

Since the Internet and the social media that come with it became massively popularized, fake news has disseminated at an unprecedented scale. It is increasingly impacting on presidential elections, celebrities, climate crisis, healthcare and many other topics.
This raising popularity of fake news may be easily observed in Fig.\ref{fig:scholar}. It presents the count of records in the \emph{Google Scholar} database appearing year by year (since 2004), related to the term "fake news".

\subsection{Overview of definitions: what is meant by fake news?}
Defining what the fake news really is poses a significant challenge. As a starting point, it is worth mentioning that Olga Tokarczuk, during her 2018 Nobel lecture \footnote{\url{https://www.nobelprize.org/prizes/literature/2018/tokarczuk/104871-lecture-english/}}, said:

\begin{quote}
\emph{"Information can be overwhelming, and its complexity and ambiguity give rise to all sorts of defense mechanisms---from denial to repression, even to escape into the simple principles of simplifying, ideological, party-line thinking. The category of fake news raises new questions about what fiction is. Readers who have been repeatedly deceived, misinformed or misled have begun to slowly acquire a specific neurotic idiosyncrasy."}
\end{quote}

There are many definitions of fake news~\cite{zhang2020overview}. In \cite{NBERw23089}, it is defined as follows: \textit{'the news articles that are intentionally and verifiably false, and could mislead readers'}. Quoting \emph{Wikipedia}, being quite more verbal and less precise, it is: \textit{'a type of yellow journalism or propaganda that consists of deliberate misinformation or hoaxes spread via traditional print and broadcast news media or online social media'}. On the other hand, in Europe, the European Union Agency for Cybersecurity (\emph{ENISA}) is using the term '\emph{online disinformation}' when talking about fake news\footnote{\url{https://www.enisa.europa.eu/publications/enisa-position-papers-and-opinions/fake-news}}. The \emph{European Commission}, on their websites, describes the fake news problem as \emph{'verifiably false or misleading information created, presented and disseminated for economic gain or to intentionally deceive the public'} \footnote{\url{https://ec.europa.eu/digital-single-market/en/tackling-online-disinformation}}. This lack of a standardized and widely accepted interpretation of the fake news term has been already mentioned in some previous papers \cite{kula2020sentiment}. It is important to remember that conspiracy theories are not always considered as fake news. In such cases, the text or images found in the considered piece of information have to be taken into account along with the motivation of the author/source.

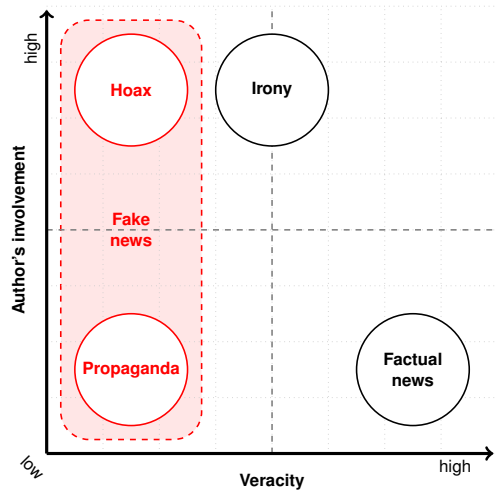
\begin{figure}
    \centering
    \resizebox{.4\textwidth}{!}{%
\begin{tikzpicture}[font=\sffamily\small,
cir/.style={draw,text width=1.8cm,minimum height=2cm, rounded corners=1cm, align=center,inner sep=.1cm, draw=black, fill=white, text=black, thick},
ar/.style={shorten >= 2pt, thick},
]

\draw[red, thick, rounded corners = .5cm, dashed, fill=red!10] (-3.75,3.75) rectangle (-1.25,-3.75);

\draw[step=1cm,black!20,thin, dotted] (-4,-4) grid (4,4);

\draw[step=6cm,gray,thick,dashed] (-4,-4) grid (4,4);

\draw[->, ar, ultra thick] (-4, -4) -- (-4,4);
\draw[->, ar, ultra thick] (-4, -4) -- (4,-4);

\node[] at (0, -4.5) {\bfseries Veracity};
\node[rotate=90] at (-4.5, 0) {\bfseries Author's involvement};

\node[cir] at (2.5,-2.5) {\textbf{Factual news}};
\node[cir, red, text=red, fill=white] at (-2.5,2.5) {\textbf{Hoax}};
\node[cir, red, text=red, fill=white] at (-2.5,-2.5) {\textbf{Propaganda}};
\node[cir] at (0,2.5) {\textbf{Irony}};

\node[text width=1cm,align=center, red] at (-2.5,0) {\textbf{Fake news}};

\node[text width=1cm,align=center, rotate=-45] at (-4.25,-4.25) {low};
\node[text width=1cm,align=center, rotate=0] at (3.25,-4.25) {high};
\node[text width=1cm,align=center, rotate=90] at (-4.25,3.25) {high};

\end{tikzpicture}}

    \caption{Fake news in the context of information}
    \label{fig:Taxonomy-2}
\end{figure}

In classification tasks it is very important to distinguish between deliberate deception (actual fake news) and irony or satire that are close to it, but completely different in the author's intention. The difference is so blurred that it is sometimes difficult even for people (especially those without a specific sense of humor), so it is a particular problem for automatic recognition systems. The definition is therefore difficult to establish, but indeed fake news is a concept related to information; therefore, we tried to position it within some other information concepts, as presented in Fig.~\ref{fig:Taxonomy-2}.

    Factual news is based on facts concerned with actual details or information rather than ideas or feelings about it.    

\subsection{Why is fake news dangerous?}
During the pandemic of Coronavirus (COVID-19) in 2020 we have had the opportunity to experience the disinformation power of fake news in all its infamous glory. The \emph{World Health Organization} called this side-phenomena the '\emph{infodemic}' -- an overwhelming quantity of overall material in social media and websites. As the representative example, one of those news items claimed that 5G mobile devices network ‘\emph{causes Coronavirus by sucking oxygen out of your lungs}’\footnote{\url{https://www.mirror.co.uk/tech/coronavirus-hoax-claims-5g-causes-21620766}}. Another group said that the virus comes from bat soup\footnote{\url{www.foreignpolicy.com/2020/01/27/dont-blame-bat-soup-for-the-wuhan-virus/}}, while others pointed labs as places where the virus was actually created as part of the conspiracy. According to the \emph{'studies'} published this way, there are also some pieces of advice on how to cure the virus -- by gargling vinegar and rosewater, vinegar and salt\footnote{\url{www.bbc.com/news/world-middle-east-51677530}} or -- as a classic for plague anxiety -- colloidal silver or garlic. 

False news may also be the grounds for political agenda, for instance during the 2016 US election. One of the misinformation examples in this campaign was the Eric Tucker's case\footnote{\url{https://www.nytimes.com/2016/11/20/business/media/how-fake-news-spreads.html}}. Tucker, finding it to be an uncommon occurrence, had photographed a big number of buses that he spotted in the city centre of Austin. Additionally, he watched the announcements concerning the demonstrations in protest at 
President-elect Donald J. Trump taking place there and arrived at the conclusion that there must have been some connection between both incidents. Tucker tweeted the photos, commenting on them that '\emph{Anti-Trump protestors in Austin today are not as organic as they seem. Here are the busses they came in. \#fakeprotests \#trump2016 \#austin}'. The original post got retweeted at least sixteen thousand times; \emph{Facebook} users shared it over 350,000 times. Later, it turned out that the buses were not involved in the protests in Austin. In fact, they were employed by \emph{Tableau Software}, a company that organised a summit for over 13 thousand people at that time. This resulted in the original post being deleted from Twitter, and instead published the picture of it labelled as '\emph{false}'. 


\section{Current initiatives worldwide to fight against disinformation}
The fake news problem is especially visible in any kind of the social media. In the online report\footnote{\url{https://techxplore.com/news/2019-12-nato-social-media.html}} NATO-affiliated researchers claimed that the social media fail to stop the online manipulation. According to the report 'Overall social media companies are experiencing significant challenges in countering the malicious use of their platforms'. First of all, the researchers were easily able to buy tens of thousands of likes, comments and views on Facebook, Twitter, YouTube and Instagram. What is more, Facebook has been recently flooded by numerous fake accounts. It claimed to disable 2.2 billion fake accounts solely in the first quarter of this year! An interesting approach to fake news detecting is to involve the community. That is why in Facebook mobile application reporting tool has become more visible lately.

This is just an example showing how serious and far-reaching the issue may be. The pervasiveness of the problem has already caused a number of fake news prevention initiatives to be developed; they are of both political and non-political character; some are local and some of them are of international scope. The following subsections will present some initiatives of considerable significance.  

\subsection{Large-scale political initiatives addressing the issue of fake news}

From the worldwide perspective, the \emph{International Grand Committee} (\emph{IGC}) \emph{on Disinformation and Fake News} can be considered as the widest present governmental initiative. The \emph{IGC} was founded by the governments of Argentina, Belgium, Brazil, Canada, France, Ireland, Latvia, Singapore, and United Kingdom. Its inaugural meeting\footnote{\url{https://www.parliament.uk/business/committees/committees-a-z/commons-select/digital-culture-media-and-sport-committee/news/declaration-internet-17-19/}} was held in the UK in November 2018 and the follow-up ones have been held in Canada (May, 2019) and Ireland (November, 2019). Elected representatives of Finland, Georgia, USA, Estonia, and Australia also attended the last meeting. In addition to general reflections about the topics under analysis, this international board specifically focused on technology and media companies, asking them for liability and accountability.
One of the conclusions of the \emph{IGC} latest meeting\footnote{\url{https://www.oireachtas.ie/en/press-centre/press-releases/20191107-update-international-grand-committee-on-disinformation-and-fake-news-proposes-moratorium-on-misleading-micro-targeted-political-ads-online/}} was that '\emph{global technology firms cannot on their own be responsible in combating harmful content, hate speech and electoral interference}'. As a result, this committee concludes that self-regulation is insufficient.

At the national level, a wide range of actions have been taken. In the case of Australia, interference in political or governmental processes has been one of the main concerns of its Government\footnote{\url{https://www.aph.gov.au/About_Parliament/Parliamentary_Departments/Parliamentary_Library/pubs/BriefingBook46p/FakeNews}}. As a result, the \emph{Electoral Integrity Assurance Taskforce} (\emph{EIAT}) was created in 2018 to handle risks (cyber interference) to the Australian election system remaining integral. Moreover, in June 2020 the Australian Communications and Media Authority published a paper\footnote{\url{https://www.acma.gov.au/sites/default/files/2020-06/Misinformation\%20and\%20news\%20quality\%20position\%20paper.pdf}} highlighting that '\emph{48\% of Australians rely on online news or social media as their main source of news, but 64\% of Australians remain concerned about what is real or fake on the internet}'. The paper discusses potentially harmful effects of fake news or disinformation to users and/or governments at different levels, providing two clear and recent examples which occurred in Australia during the first semester of 2020, such as the bushfire season and the COVID-19 pandemic. In general, two responses to misinformation are pointed out. One that considers international regulatory responses and another one coming from online platforms in terms of how they tackle misconducting users as well as how they address problematic content.

Similarly, in October 2019 the \emph{US Department of Homeland Security} (\emph{DHS}) published a report on \emph{Combatting Targeted Disinformation Campaigns}\footnote{\url{https://www.dhs.gov/sites/default/files/publications/ia/ia_combatting-targeted-disinformation-campaigns.pdf}}. In this report, the \emph{DHS} highlights how easy it is nowadays to spread false news through online sources and how '\emph{disinformation campaigns should be viewed as a whole-of-society problem requiring action by government stakeholders, commercial entities, media organizations, and other segments of civil society}'. This report points out the growth on disinformation campaigns since the 2016 US presidential election; at the same time that US and world-wide nations were becoming more aware concerning the potential damage of these campaigns to economy, politics and society in general. Furthermore, better and wider actions are nowadays carried out in real-time, i.e. while the disinformation campaign is ongoing, compared to the first years (until 2018) where most of '\emph{the work on disinformation campaigns was post-mortem, i.e. after the campaign had nearly run its course}'. In this sense, the report summarizes several recommendations to combat disinformation campaigns such as '\emph{government legislation, funding and support of research efforts that bridge the commercial and academic sectors (e.g., development of technical tools), sharing and analysing information between public and private entities, providing media literacy resources to users and enhancing the transparency of content distributors, building societal resilience and encouraging the adoption of healthy skepticism}'. In any case, the importance of '\emph{private and public sector cooperation to address targeted disinformation campaigns on the next five years}' is highlighted.

In Europe, the \emph{EU Commission} has recently updated (July, 2020) a previous report providing a clear set of actions, which are easy to understand and/or apply, in order to fight against fake news and online disinformation\footnote{\url{https://ec.europa.eu/digital-single-market/en/tackling-online-disinformation}}$^,$\footnote{\url{https://ec.europa.eu/newsroom/dae/document.cfm?doc_id=56166}}. The action plan consists of the four following pillars:
\begin{enumerate}
    \item \emph{Improving the capabilities of Union institutions to detect, analyse and expose disinformation}. This action implies better communication and coordination among the EU Member States and their institutions. In principle, it aims to provide EU members with '\emph{specialised experts in data mining and analysis to gather and process all the relevant data}'. Moreover, it refers to '\emph{contracting media monitoring services as crucial in order to cover a wider range of sources and languages}'. Additionally, it also highlights the need to '\emph{invest in developing analytical tools which may help to mine, organise and aggregate vast amounts of digital data}'.
    \item \emph{Stronger cooperation and joint responses to threats}. Since the most significant is the time right after the publishing of the false news, this action aims to have a '\emph{Rapid Alert System to provide alerts on disinformation campaign in real-time}'. For this purpose, each EU Member State should '\emph{designate contact points which would share alerts and ensure coordination without prejudice to existing competences and/or national laws}'.
    \item \emph{Enhancing collaboration with online platforms and industry to tackle disinformation}. This action aims to mobilise and provide with an active role to the private sector. Disinformation is most often released in large online platforms owned by the private sector. Therefore, they should be able to '\emph{close down fake accounts active on their service, identify automated bots and label them accordingly, and collaborate with the national audio-visual regulators and independent fact-checkers and researchers to detect and flag disinformation campaigns}'.
    \item \emph{Raising awareness and improve societal resilience}. This action aims to increase public awareness and resilience by activities related to '\emph{media literacy in order to empower EU citizens to better identify and deal with disinformation}'. In this sense, the development of critical thinking and the use of independent fact-checkers are highlighted to play a key role in '\emph{providing new and more efficient tools to the society in order to understand and combat online disinformation}'.
\end{enumerate}

Overall, any activity or action proposed in the above-mentioned cases (\emph{IGC}, Australia, US and EU) could be understood from three different angles: technological, legislative and educative. For instance -- technology-wise -- the four pillars in the set of actions proposed by the EU Commission are mentioning the use of several kinds of tools (analytical, fact-checkers, etc.), arising as the key component to be considered in current and/or future initiatives. However, from the legislation point of view, pillars 1 and 2 are also showing the need to have a normative framework which facilitates coordination and communication among different countries. Finally -- education-wise -- pillars 3 and 4 strengthen the importance of media literacy and the development of critical thinking in society. Similarly, activities, actions and recommendations provided by the \emph{IGC}, the Australian government and the \emph{DHS} can be directly linked to these three different concepts (technology, legislation and education). Furthermore, the \emph{BBC} has an available collection of tools, highlights and global media education\footnote{\url{https://www.bbc.co.uk/academy/en/collections/fake-news}} which could be directly linked to these angles previously described, thus supporting this division into three categories. 

The U.S. Department of Defence started to invest about \$70 M in order to deploy military technology to detect fake contents as they impact on national security\footnote{\url{https://www.cbc.ca/news/technology/fighting-fake-images-military-1.4905775}}. This \emph{Media Forensics Program} at the \emph{Defence Advanced Research Projects Agency} (\emph{DARPA}) started 4 years ago. 

Some efforts are being devoted to the technological component from public bodies. That is the case of the Australian \emph{EIAT}, which was created to provide the \emph{Australian Electoral Commission} with ‘\emph{technical advice and expertise}’ concerning potential digital disturbance of electoral processes\footnote{\url{https://parlinfo.aph.gov.au/parlInfo/search/display/display.w3p;query=Id\%3A\%22media\%2Fpressclp\%2F6016585\%22}}. The \emph{Australian Cyber Security Centre} was in charge of providing this technological advice, among other governmental institutions.

In the case of Europe, a high-level group of experts (\emph{HLGE}) was appointed in 2018 by the European Commission to give advice on this topic. Although these experts were not against regulating in some cases, they proposed~\cite{EU-report} to mainly take non-regulatory and specific-purpose actions. The focus of the proposal was the collaboration between different stakeholders to support digital media companies in combating disinformation. On the contrary, the \emph{Standing Committee on Access to Information, Privacy and Ethics} (\emph{SCAIPE}) established by the Parliament of Canada suggested in 2019 to impose legal restrictions to media companies in order to be more transparent and force them to remove illegal contents~\cite{canadaReport}. Similarly, the \emph{Digital, Culture, Media and Sport Committee} (\emph{DCMSC}) created by the Parliament of the United Kingdom strongly encouraged to take legal actions~\cite{UKReport}. More precisely, it proposed a mandatory ethical code for media enterprises, to be controlled by an independent body, and forcing these companies to remove those contents to be considered potentially dangerous and coming from proven sources of disinformation. Furthermore, the \emph{DCMSC} suggested to modify laws regarding electoral communications to ensure their transparency in online media. 

Regarding electoral processes, it is worth mentioning the Australian initiative; unprecedented offences by foreign interference have been added to the Commonwealth Criminal Code by means of the \emph{National Security Legislation Amendment (Espionage and Foreign Interference) Act 2018}\footnote{\url{https://www.legislation.gov.au/Details/C2018C00506}}. It defines these foreign offences as the ones that '\emph{influence a political or governmental process of the Commonwealth or a State or Territory or influence the exercise (whether or not in Australia) of an Australian democratic or political right or duty}'.

In the case of India, the Government issued a notice to \emph{Whatsapp} in 2018, because at least 18 people were killed in separate incidents that year after false information was shared through this app\footnote{\url{https://www.cbc.ca/news/world/india-child-kidnap-abduction-video-rumours-killings-1.4737041}}. The Minister of Electronics and Information Technology stated that the Indian Government '\emph{was committed to freedom of speech and privacy as enshrined in the constitution of India}'. As a result, the posts published to social networks are not subject to governmental regulations. He claimed that '\emph{these social media have also to follow Article 19(2) of the Constitution and ensure that their platforms are not used to commit and provoke terrorism, extremism, violence and crime}'\footnote{\url{https://www.thehindu.com/news/national/fake-news-safety-measures-by-whatsapp-inadequate-says-ravi-shankar-prasad/article24521167.ece}}. However, India's Government is working on a new IT Act in order to deploy a stronger framework to deal with cybercrimes\footnote{\url{https://www.thehindu.com/business/Industry/centre-to-revamp-it-act/article30925140.ece}}.

Lastly, it should be mentioned that the third (latest) meeting of the \emph{IGC} was aimed at advancing international collaboration in the regulation of fake news and disinformation. In this sense, experts highlighted that there are conflicting principles regarding the regulation of the internet. This includes the protection of freedom of speech (in accordance with national laws), while simultaneously combating disinformation. Thus, this still is an open challenge.

Finally, from the education perspective, it is worth mentioning that the \emph{EU-HLGE} recommended implementing wide education programs on media and information, in order to educate not only professional users of media platforms but public users in general terms. Similar recommendations were pointed out by the Canadian \emph{SCAIPE}, focusing on awareness-raising campaigns and literacy programs for the whole society.

\subsection{Other noteworthy initiatives and solutions}

It should also be mentioned that recently some systematic social activity battling with misinformation has appeared and now it is getting more intense. For instance, there is a group of volunteers called Lithuanian '\emph{elves}'. Their main aim is to beat the \emph{Kremlin} propaganda. They scan the social media (\emph{Instagram, Facebook, Twitter}) and report any found fake information on their daily basis.

From the technological perspective, it is worth highlighting that numerous online tools have been developed for misinformation detection. Some available approaches were presented in~\cite{gielczyk2019evaluation}. 
This technological development to combat disinformation is led by tech/media companies. This is the case of \emph{Facebook}, that quite recently (May 2020) informed\footnote{\url{https://spectrum.ieee.org/view-from-the-valley/artificial-intelligence/machine-learning/how-facebook-is-using-ai-to-fight-covid19-misinformation}} that it is combating fake news by means of its \emph{Multimodal Content Analysis Tools}, in addition to 35 thousand human supervisors. This AI-driven set of tools is been applied to identify fake or abusive contents related to Coronavirus. The image-processing system extracts objects that are known to violate its policy. Then the objects are stored and searched in new ads published by users. \emph{Facebook} claims that this solution, based on supervised classifiers, does not suffer from the limitations of similar tools when facing images created by common adversarial modification techniques. 

In the \emph{BlackHat Europe 2018} event that was held in London, the \emph{Symantec Corporation} displayed its demo of a deepfake detector\footnote{\url{https://i.blackhat.com/eu-18/Thu-Dec-6/eu-18-Thaware-Agnihotri-AI-Gone-Rogue-Exterminating-Deep-Fakes-Before-They-Cause-Menace.pdf}}. In 2019, Facebook put 10 million dollars\footnote{\url{https://www.reuters.com/article/us-facebook-microsoft-deepfakes/facebook-microsoft-launch-contest-to-detect-deepfake-videos-idUSKCN1VQ2T5}} into the \emph{Deepfake Detection Challenge}~\cite{DFDC2020} aimed at measuring progress on the available technology to detect deepfakes. The best model (in terms of precision metric for published data) that won this contest, performed quite poorly (65.18\% precision) when validated with new data. This means that it still is an open challenge and a great research effort is still required to get robust fake detection technology. In addition to these huge companies, some other startups are developing anti-fake technologies, such as the \emph{DeepTrace} based in Netherlands. This company aims at building the '\emph{antivirus for deepfakes}'\footnote{\url{https://spectrum.ieee.org/tech-talk/artificial-intelligence/machine-learning/will-deepfakes-detection-be-ready-for-2020}}. 

Some other technological projects are being run at present time; the \emph{AI Foundation} raised 10 million dollars to develop the \emph{Guardian AI} technologies, a set of tools comprising \emph{Reality Defender}\footnote{\url{https://aifoundation.com/responsibility/}}. This intelligent software is intended to support users in identifying fake contents while consuming digital resources (such as web browsing). No further technical details are available yet.

\section{A systematic overview of ML approaches for fake news detection}

A comprehensive, critical analysis of previous ML-based approaches to false news detection is presented in the following part of the paper. As already mentioned in Section 1, and as graphically presented in Fig. \ref{fig:approaches}, these methods can analyze different types of digital content. According to that, in this section we will overview the methods for (\emph{i}) text-based and Natural Language Processing (NLP) analysis, (\emph{ii}) reputation analysis, (\emph{iii}) network analysis, and (\emph{iv}) image-manipulation recognition.

\subsection{Text analysis}
Intuitively, the most obvious approach to automatically recognizing fake news is NLP. Despite the fact that the social context of the message conveyed in electronic media is a very important factor, the basic source of information necessary to build a reliable pattern recognition system is the extraction of features directly from the content of the analyzed article. Several main trends may be distinguished within the works carried out in this area. Theoretically, the simplest is the analysis of text representation without linguistic context, most often in the form of \emph{bag-of-words} (first mentioned in 1954 by Harris \cite{harris1954distributional}) or \emph{N-grams}, but also the analysis of psycholinguistic factors, syntactic and semantic analysis are commonly used.

\subsubsection{NLP-based data representation}
As it is rightly pointed out by Saquete et al. \cite{Saquete2020}, each of the subtasks distinguished within the domain of detecting false news is based on the tools offered by NLP. Basic solutions are built on \emph{bag-of-words}, which counts the occurrences of particular terms within the text. A slightly more sophisticated development of this idea are \emph{N-grams}, which tokenize not individual words, but their sequences. The range of the definition allows to define \emph{bag-of-words} as \emph{N-grams} with $n$ equal to one, making them \emph{unigrams}. It has also to be noted that the sheer number of \emph{N-grams} in each document is strongly dependent on its length and for the needs of the construction of pattern recognition systems it should be normalized to the document (\emph{Term frequency} : TF \cite{luhn1957statistical}) or to the set of documents used in the learning process (\emph{Term frequency - inverse document frequency : TF-IDF} \cite{jones1972statistical}).

Despite the simplicity and age of these solutions, they are successfully utilized in solving the problem of detecting false news. A good example of such application is the work of Hassan et al. \cite{Hassan2020291}, comparing the effectiveness of \emph{Twitter} disinformation classification using five base classifiers (\emph{Lin-SVM, RF, LR, NB and KNN}), comparing them with methods of TF and TF-IDF attribute extraction with \emph{N-grams} of various lengths. On the basis of the PHEME \cite{zubiaga2016pheme} dataset, they showed the effectiveness of methods using simultaneously different lengths of word sequences, combining \emph{unigrams} with \emph{bigrams} in the extraction. A similar approach is a common starting point for most analyzes \cite{bharadwaj2019fake, Wynne2019}, enabling further suggestions for considerations, through the in-depth review of base classifiers \cite{kaur2020automating} or the use of ensemble methods \cite{Ksieniewicz:2019}.

Other interesting trends within this type of analysis include the detection of unusual tokens, for example text elements, insistently repeated denials and curses, or question marks, emoticons and multiple exclamation marks, that are most often rejected at the stage of preprocessing \cite{horne2017just, Castillo:2011:ICT:1963405.1963500, Telang2019}. As in many other application fields, deep neural networks are a promising branch of Artificial Intelligence. Hence, they are also playing a role here, being a popular alternative to classic models \cite{Kong2020102}.

\subsubsection{Psycholinguistic features}

The psycholinguistic analysis of texts published on the Internet is particularly difficult due to the limitation of messages to their verbal part only and the peculiar characteristics of such documents. Existing and widely cited studies \cite{zhou2008following,zhang2016online} allow us to conclude that the messages that try to mislead us are characterized, for example, by an enormous length of some sentences they contain along with their lexical limitation, increased repetition of key theses or a reduced formality of the language used. These are often extractable and measurable factors that can be more or less successfully used in the design of the pattern recognition system.

An interesting work giving a proper overview on psycholinguistic data extraction is the detection of character assassination attempts by troll actions, performed by El Marouf et al. \cite{Marouf2019}. Six different feature extraction tools were used in the construction of the dataset for supervised learning. The first is \emph{Linguistic Inquiry and Word Count} (\emph{LIWC}) \cite{pennebaker2001linguistic}, which in its latest version allows to obtain 93 individual characteristics of the analyzed text, including both simple measurements, like the typical word count within a given phrase and complex analysis of the grammar used or even the description of the author's emotions or the cognitive processes performed by them. It is a tool that has been widely used for many years in a variety of problems (\cite{ott2011finding}, \cite{Robinson2013}, \cite{huang2012development}, \cite{delPilarSalasZrate2014}), fitting well for detecting fake news. Notwithstanding, in~\cite{fariasirony} authors presented a sentiment analysis proposal that classifies the sample text as irony or not.

Another tool are \emph{POS} Tags, assigning individual words to \emph{Parts-of-speech} and returning their percentage share in the document \cite{Stoick2019}. The basic information about the grammar of the text and the presumed emotions of the author obtained in this way are supplemented with the knowledge acquired from \emph{SlangNet} \cite{dhuliawala2016slangnet}, \emph{Colloquial WordNet} \cite{mccrae2017colloquial}, \emph{SentiWordNet} \cite{baccianella2010sentiwordnet} and \emph{SentiStrength} \cite{thelwall2017heart}, returning, in turn, information about slang and colloquial expressions, indicating the author's sentiment and defining it as positive or negative. The system proposed by the authors, using 19 of the available attributes and \emph{Multinomial Naive Bayes} as a base classifier, allowed to obtain a 90\% score in the \emph{F-score} metric.

An extremely interesting trend in this type of feature extraction is the use of behavioral information that is not directly contained in the entered text, but can be obtained by analyzing the way it was entered \cite{ahmed2013biometric, ahmed2018detecting}.


\subsubsection{Syntax-based}

The aforementioned methods of obtaining useful information from the text were based on the analysis of the word sequences present in it or the recognition of the author's emotions hidden in the words. However, a full analysis of \emph{natural language} requires an additional aspect in the form of processing the syntactic structure of expressed sentences. Ideas expressed in simple data representation methods, such as \emph{N-grams}, were developed to \emph{Probability Context Free Grammars} \cite{stolcke1994precise}, building distributed trees describing the syntactic structure of the text. 

In syntactic analysis, we do not have to limit ourselves to the structure of the sentence itself, and in the case of social networks, extraction can take place by building the structure of the whole discussion, as Kumar and Carley show \cite{Kumar20205047}. Extraction of this type \cite{tai2015improved} seems to be one of the most promising tools in the fight against fake news propagation \cite{zubiaga2016stance}.

\subsubsection{Non-linguistic methods}

The fake news classification is not limited to the linguistic analysis of documents. The studies analysing different kinds of attributes which may be of use in the same setting \cite{Shu:2017} are interesting. Amidst the typical methods that the study comprises, there are the analyses of the creator and reader of the message, as well as the contents of the document and its positioning within social media outlets being verified \cite{Zhang:2019}. Another method which shows promise is analysing images; this approach concerns fake news in the form of video material \cite{choras2018pattern}. Similarly, the study by \cite{conroy2015automatic} is evenly thought-provoking; it proposes to divide the methods of linguistic and social analyses. The former group of models encompasses the semantic, rhetorical, discourse and simple probabilistic recognition ones. The latter set comprises analysing how the person conveying the message behaves in social media and what context their entries are building. Then, \cite{Castillo:2011:ICT:1963405.1963500} has based the design of the recognition models on the behavior of the authors, where the background of the post (both the posted and forwarded ones) does depend on their bodies, and at the same time refers to other texts. Diverse representations of data were analysed by \cite{Ferrara:2016}, whilst \cite{Afroz:2012} has examined various variants of stylometric metrics.

Numerous issues relating to fake news detection have been studied by \cite{sharma2019combating} and indicated that it is possible to apply the \emph{Scientific Content Analysis} (SCAN) approach in order to tackle the matter. In \cite{Jin:2017}, an effective method was advanced which aims at verifying the news automatically. This approach would depart from analysing texts and head for image data. On the other hand, \cite{Zhang:2019} suggests performing analyses of the posts from social networks within the context of data streaming, arguing that this approach addresses their dynamic character. Support Vector Machine (SVM) has been utilised by \cite{horne2017just} in order to recognize which messages are genuine, false or of satirical nature. A similar type of classifier has been employed by \cite{Chen:2011} as well; in their work, semantic analysis as well as behavioural feature descriptors are to uncover the media entries which may be false.

The comparison was made by \cite{Gravanis:2019} so as to assess a number of classification methods which base on linguistic features. According to its outcomes, successful detection of false information can base on the already familiar classifier models (particularly ensembles). In \cite{Bondielli:2019}, it has been indicated that the issue of detecting false information is generally limited to the classification task, although anomaly detection and clustering approaches might be utilised for it. Lastly, in \cite{Atodiresei:2018}, the NLP tools-based method was proposed to be applied for analysing Twitter posts. According to this approach, each post was assigned credibility values owing to the fact that the researchers viewed this issue as a regression task. 

\subsection{Reputation analysis}
The reputation\footnote{\url{https://www.definitions.net/definition/REPUTATION}} of an individual, a social group, a company or even a location, is understood as the estimation of it; usually it results from the determined criteria which influence social assessment. Within a natural society, reputation is the mechanism of social control, characterised by high efficiency, ubiquity and spontaneity. Social, management and technological sciences aim at investigating this matter. It must be acknowledged that reputation influences communities, markets, companies, and institutions alike; in other words, it has an influence over both the competitive and cooperative contexts. Its influence may extend as far as the relations between the whole countries; the idea's importance is appreciated in politics, business, education, online networks, and diverse, innumerable domains. Thus, reputation can be regarded to be reflecting the identity of a given social entity. 

In technological sciences, the reputation of a product or a site often needs to be measured. Therefore a reputation score, which represents it in a numeric manner, is computed. The calculation may be performed by means of a centralized reputation server or distributively, by means of local or global trust metrics~\cite{buford2009p2p}. This evaluation may be of use when supporting entities in their decisions concerning if they should rely on someone or buy a given product, or not.

The general concept behind reputation systems is allowing the entities evaluate one another or assess an object of their interest (like articles, publishers, etc.); then subsequently utilize the collected evaluations to obtain trust or reputation scores, concerning the sources and items within the system. In order for systems to act in this way, reputation analysis techniques are used, which actually support the ability to establish in an automatic manner the way various keywords, phrases, themes or contents created by users are analysed amongst the mentions of diverse origin. More specifically, in the news industry, there are two kinds of sources of reputation for an article/publisher~\cite{xu2019detecting}; reputation from content and reviews/feedback on the one hand, and reputation from IP or domain on the other one.

Given the technological advances, all types of data can be collected: type of comments, their scope, keywords, etc. Especially in the news industry, there are a few key characteristics that can differentiate a trustworthy article from a fake one. A common characteristic is the anonymity fake news publishers choose behind their domain. The results from the survey in~\cite{xu2019detecting} showed that whenever the person publishing contents wishes to protect their anonymity, the online who-is information will indicate the proxy as the organization that registered it. On the other hand, the renowned, widespread newspapers usually prefer to register under their actual company names. Another indicative feature for fake news is the duration of time that publishers spend on the websites with the intention of disseminating false information. Often, it is rather brief in comparison to the real news publishers. Domain popularity is also a good indicator regarding the reputation, as it measures the views of the website gets every day, per a visiting person. It seems logical that a well-know website features a greater number of views per person, as they have the tendency to devote more time to surfing the contents and looking at the various sub pages. It has been indicated in~\cite{xu2019detecting} that the domains of the trustworthy web pages which post genuine information are far more popular than the ones which disseminate untruth. The reason for this is that normally the majority of web pages that publish false news either stop publishing news very soon, or the readers spend much less time browsing those sites.

So, reputation scoring ranks fake news based on suspicious domain names, IP addresses and review/feedback by the readers by providing a measure that indicates whether the specific website is high or low on reputation. Different techniques for reputation scoring have been proposed in the literature. ~\cite{hegli2013system} describes the application of the \emph{Maximum Entropy Discrimination} (MED) classifier. It is utilized to score the reputation of web pages. It is done on the basis of the information which creates a reputation vector. It includes multiple factors for the online resource, such as the state in which the domain was registered, the place where the service is hosted, the place of an internet protocol address block and when the domain was registered. It also considers the popularity level, IP address, how many hosts there are, top-tier domain, a number of run-time behaviours, \emph{JavaScript} block count, the number of images, immediate redirect, and response latency. The paper by~\cite{antonakakis2010building} also relates to the issue. The scientists have created the \emph{Notos} reputation system which makes use of the unique DNS characteristics in filtering out the malevolent domains on the basis of their past engagement in harmful or genuine online services. For every domain, the authors utilised clustering analysis of network-based, zone-based and evidence-based features, so as to obtain the reputation score. As the majority of the methods do not perform the scoring online, so it may use processing-intensive approaches. The assessment that used real-world data, including the traffic from vast ISP networks, has proved that \emph{Notos}'s accuracy in recognizing emerging, malevolent domains in the DNS query traffic that was monitored was indeed very high, with the true positive score of 96.8\% and false positive one - 0.38\%.

According to the above mentioned, a reputation analysis system is actually a cross-checking system that examines a set of trustworthy databases of blacklists in an automatic manner, 
usually employing ML techniques that recognize malicious IP addresses and domains based on their reputation scores.
More specifically, in~\cite{lison2017neural}, a ML model relying on a deep neural architecture which was trained using a big passive DNS database is presented. \emph{Mnemonic}\footnote{\url{https://www.mnemonic.no/}} supplied the passive DNS data utilised for the sake of the paper. The raw dataset consisted of 567 million aggregated DNS queries, gathered in the span of almost half a decade. The variables which define every entry are as follows: the type of a record, a recorded query, the response to it, a \emph{Time-to-Live} (TTL) value for the query-answer pair and a timestamp for when the pair occurred at first, as well as the total number of instances when the pair occurred, within a given period. The method is capable of pinpointing 95\% of the suspicious hosts, the false positive rate being 1:1000. Nevertheless, the amount of time required to train turned out to be exceptionally high because of the vast amount of the data needed for it; the delay information has not been assessed. 

\emph{Segugio}, an innovative defense system based on behaviour, is introduced in~\cite{rahbarinia2015segugio}. It makes it possible to track the appearance of newly-appearing, malware-control domain names within huge ISP networks in an efficient manner, with the true positive rate (TP) reaching 85\%, and false positive rate (FP) lower than 0.1\%. Nevertheless, both TP and FP have been counted on the basis of 53 new domains; proving the correctness based on such a little set may be a challenging task.

Lastly, in~\cite{tang2006fast}, an innovative novel granular SVM is presented, namely a boundary alignment algorithm (GSVM-BA), which repeatedly eliminates the positive support vectors from the dataset used for training, in order to find the optimum decision boundary. To accomplish it, two groups of feature vectors are extracted from the data; they are called the breadth and spectral vectors.

\subsection{Network data analysis}
Network analysis refers to the Network theory, which studies graphs, being a representation of either symmetric or asymmetric relations between discrete objects. This theory has been of use in various fields including statistical physics, particle physics, computer science, electrical engineering, biology, economics, and others. The possible applications of the theory comprise the World Wide Web (WWW), Internet, as well as logistical, social, epistemological and gene regulatory networks, etc. In computer/network sciences, the network theory belongs to graph theory. This means that one may define a network as a graph, where nodes and edges have their attributes (like names).

Network-based detection of false news applies the data on the social context uncovered in news propagation. Generally speaking, it examines two kinds of networks, namely the homogeneous and heterogeneous ones. Homogeneous networks (such as Friendship, Diffusion, and Credibility networks, etc.) contain one kind of nodes and edges. For instance, in credibility networks, people present their points of view regarding the original news items by means of social media entries. In them, they might either share the same opinions (which thus support one another), or conflicting opinions (this is turn may lower their credibility scores). If one were to model the aforementioned relations, the credibility network may be applied to assess the level of veracity of the news pieces, by means of the credibility scores of every particular social network entry (related to the news item) being leveraged. On the other hand, heterogeneous networks are characterised by having numerous kinds of nodes or edges. The main benefits they bring is the capability of representing and encoding the data and relations from various positions. Some well-known networks used for detecting false information are Knowledge, Stance, and Interaction Networks. The first type incorporates linked open data, like DBdata and Google Relation Extraction Corpus (GREC), as a heterogeneous network topology. When inspecting for fact by means of a knowledge graph, it is possible to verify if the contents of the news items may be gathered from the facts that are present in the knowledge networks, whilst Stances (viewpoints) represent people's attitudes to the information, i.e. in favour, conflicting, etc, ~\cite{shu2019studying}.

In detecting false news, network analysis is performed in order to evaluate the truth value of the news item; one may formalize it as a classification issue which needs obtaining relevant features and building models. As part of feature extraction, the differential qualities of information items become captured in order to create efficient representations; on the basis of them, several models are created, for learning at transforming the features. Contemporary advancements in network representation learning, e.g. network embedding and deep neural networks, let one apprehend the features of news items in an enhanced manner, from supplementary data like friendship networks, temporal user engagements, and interaction networks. Moreover, knowledge networks as secondary data may make it easier to challenge the truthfulness of the news pieces by means of network pairing operations, including path finding and optimizing the flow.

\par Across network levels, the data from the social media concerning propagating the news and its spreaders has not been examined to a significant degree, yet. In addition to this, it has not been much used in an explainable manner for detecting false information, too. The authors of ~\cite{zhou2019network} have suggested a network-based pattern-driven model for detecting false information; it proved robust against the news items being manipulated by malicious actors. The model makes use of patterns in disseminating fake data within social networks, as it turns out that, in comparison with the true ones, false news is able to 
(\emph{i}) disseminate further and (\emph{ii}) engage a larger number of spreaders, where they oftentimes prove to be (\emph{iii}) more fully absorbed in the news and (\emph{iv}) more closely linked within the network. The characteristics which represent the aforementioned patterns have been developed at several network levels (that is, node-, ego-, triad-, community-, and network-level), that may be utilised in a ML supervised-learning framework for false news detection. The patterns involved in the mentioned study regarding fake news concern the spreading of the news items, the ones responsible for doing it, and the relations among those spreaders. Another example of network analysis in the false news detection may be found in~\cite{shu2017exploiting}, where a framework is proposed which uses a tri-relationship model (TriFN) amongst the news article, spreader and publisher. For such a network, a hybrid framework is composed of three major parts which contribute to detecting false news: (\emph{i}) entity embedding and representation, (\emph{ii}) relation modeling, and (\emph{iii}) semi-supervised learning. The actual model possesses four meaningful parameters. $\alpha$ and $\beta$ manage the inputs from social relationship and user-news relations. $\gamma$ manages the input of publisher partisan and $\eta$ controls the input provided by semi-supervised classifier. According to ~\cite{shu2017exploiting}, TriFN is able to perform well whilst detecting, at the initial stages of disseminating news.

\subsection{Image based analysis and detection of image manipulations}

In the last decade, digital images have thoroughly replaced conventional photographs. Currently it is possible to take a photo using not only cameras but also smartphones, tablets, smart watches and even eye glasses. Thus, thousands of billions of digital photos are taken annually. The immense popularity of image information fosters the development of the tools for editing it, for instance \emph{Photoshop} or \emph{Affinity Photo}. The software lets users manipulate real-life pictures, from low-level (adjusting the brightness in a photo) to high-level semantic contents (replacing an element of a family photograph). Nonetheless, the possibilities provided by the photo manipulation tools may be seen as a double-edged sword. It enables making the pictures more pleasing to the eye, and also inspires users in their expressing and sharing their visions on visual arts; however, the contents of the photo may be forged more easily, with no visible hints left. Thus, it makes it easier to spread fake news. Hence, with the passage of time, several scientists have developed the methods for detecting photo tampering; the techniques concentrate on copy-move forgeries, splice forgeries, inpainting, image-wise adjustments (such as resizing, histogram equalization, cropping, etc.) and other ones.

So far, numerous researchers have presented some approaches for detecting fake news and image tampering. In~\cite{bondi2017tampering}, the authors proposed an algorithm which is able to recognise if the picture has been altered and where, taking advantage of the characteristic footprints that various camera models leave in the images. Its way of working is based on the fact that all the pixels of the pictures that have not been tampered with ought to appear as if they had been taken by means of a single device. Otherwise, if the image has been composed of multiple pictures, then the footprints left by several devices may be found. In the presented algorithm, a \emph{Convolutional Neural Network} (CNN) is exploited for obtaining the characteristics which are typical of the specific camera model, from the studied image. The same algorithmic was also used in~\cite{zhang2020dense}.

Authors in \cite{james2019image} used \emph{Structural Similarity Index Measure} (SSIM) instead of more classical approaches to modification detection techniques such as: \emph{mean squared error} (MSE) and \emph{peak signal to noise ratio} (PSNR). Moreover, they moved the whole solution to cloud computing that is claimed to provide security, the information being deployed quicker, as well as the data being more accessible and usable. 

The technique presented in \cite{kanwal2020digital} uses the chrominance of the \emph{YCbCr} colour space; it is considered to be able to detect the distortion resulting from forgery more efficielty than all the further colour components. When extracting features, selected channel gets segmented into blocks which overlap. Thus, \emph{Otsu-based Enhanced Local Ternary Pattern} (OELTP) has been introduced; it is an innovative visual descriptor aimed at extracting features from the blocks. It extends the \emph{Enhanced Local Ternary Pattern} (ELTP) which recognises the neighbourhood on a range of three values (-1, 0, 1). Subsequently, the energy of OELTP features gets assessed in order to decrease the dimensionality of features. Then, the sorted characteristics are used for training the SVM classifier. Lastly, the image is labelled, either as genuine or manipulated.

The colour space \emph{YCbCr} was also used in \cite{dua2020detection}. The presented approach takes advantage of the fact that during when being merged, at least two pictures get engaged in copying and pasting. In case of tampering with JPEG images, the forgery might not follow the same pattern; a piece of an uncompressed picture may be pasted into a compressed JPEG file or the other way round. Such manipulated images being re-saved in JPEG format with different quality factors can possibly result in the emergence of double quantization artefacts in DTC coefficients.

According to the method proposed in \cite{jaiswal2020technique}, the input file becomes pre-processed by converting its colour space from RGB to grey level; following that, the image features (\emph{Histogram of Oriented Gradient} (HOG), \emph{Discrete Wavelet Transform} (DWT) and \emph{Local Binary Patterns} (LBP)) get distilled from the grey level colour space. This fitting group of features is merged to create a feature vector. The \emph{Logistic Regression} classifier is utilised to construct a model and to discriminate a manipulated, authenticated picture. The suggested technique enhances the correctness of detection by applying combined spacial features, that is the spatial- and frequency-based ones.

Two types of features were also used in \cite{jothi2020tampering}. In this work, the authors decided to convert the source image into grey scale and use \emph{Haar Wavelet Transform}. Then, the vector features are calculated using HOG and \emph{Local Binary Patterns Variance}. The classification step is founded upon the Euclidean distance.

\par Within \cite{bilal2019single}, \emph{Speeded Up Robust Features} (SURF) and \emph{Binary Robust Invariant Scalable Keypoints} (BRISK) descriptors were used in order to reveal and pinpoint single and multiple copy–move forgeries. The features of SURF prove robust against diverse post-processing attacks, like rotation, blurring and additive noise. Nevertheless, it seems that features of the BRISK are equally as robust in relation to detecting the scale-invariant forged regions, along with the poorly localized keypoints of the objects within the forged image. 

However, the \emph{state-of-the-art} techniques deal not only with copy-move forgeries, but with other types of modifications, too. The paper \cite{suryawanshi2019detection} presents the contrast enhancement detection based on numerical measures of images. The proposed method includes division in non-overlapping blocks and then, mean, variance, skewness and kurtosis of block calculation. The method also uses DWT coefficients and SVM for original/tampered classification. Whereas, in \cite{guo2018fake} authors claimed that real and fake colorized images can differ in \emph{Hue} and \emph{Saturation} channels of \emph{Hue-Saturation-Value} (HSV) colour space. Thus, the presented approach using histogram equalisation and some other statistical features enables authenticity verification in the colorizing domain.

Overview of all the tools of feature extraction from fake news mentioned in this section is presented in Table \ref{tab:extractors}. 

\begin{table}[ht]
\caption{Overview of ML extractors for fake news detection.}\vspace{.5em}
    \centering
        \begin{tabularx}{\textwidth}{lp{2.5cm}L}
        \toprule
        \bfseries \textsc{category} & \bfseries \textsc{extractor} & \bfseries \textsc{context of extraction}\\
        \midrule
        
        \textsc{nlp} & \emph{N-grams} & Primitive tool of tokenization.\\
        &\emph{TF} & Normalization of tokens to the documents.\\
        &\emph{TF-IDF} & Normalization of tokens to the corpora.\\[1em]
        
        \textsc{psycholinguistic} & \emph{LIWC} & Collection of 93 individual characteristics from word counts to grammar analysis.\\
        &\emph{POS Tags} & Assignment of words to parts of speech.\\
        &\emph{SlangNet} & Various models of recognition of slang and colloquial expressions. \\
        &\emph{ColloquialWordNet}\\
        &\emph{SentiWordNet} \\
        &\emph{SentiStrength} \\[1em]
        \textsc{syntax} & \emph{PCFG} & Construction of distributed trees describing dynamic structure of text.\\
        & \emph{Tree LSTMs} & Application of Long Short-Term Memory networks to analysis of content distribution.\\[1em]
        
        \textsc{non-linguistic} & \emph{SCAN} & Scientific content analysis.\\[1em]

        \textsc{reputation} & \emph{MED} & Reputation vector of a publisher.\\
        & \emph{Notos} & Reputation vector from DNS characteristic.\\
        & \emph{Segugio} & Tracking the appearance of new, malware-control domain names within huge ISP networks.\\[1em]
        
        \textsc{image} & \emph{CNNs} & Primitive tool of feature extraction of digital signals.\\
        & \emph{SSIM} & Modification detection metric.\\
        & \emph{OELTP} & Visual block descriptor.\\
        & \emph{HOG, DWT, LBT} & Primitive tools of feature extraction.\\
        & \emph{SURF, BRISK} & Detection of copy-move forgeries.\\
        \bottomrule
        \end{tabularx}
\label{tab:extractors}
\end{table}

\section{Research interest and popular datasets used for detecting fake news}

\subsection{Is fake news an important issue for the research society?}
\label{chap:researchSociety}
\begin{figure}[!ht]
    \centering
    \includegraphics[width=.8\textwidth]{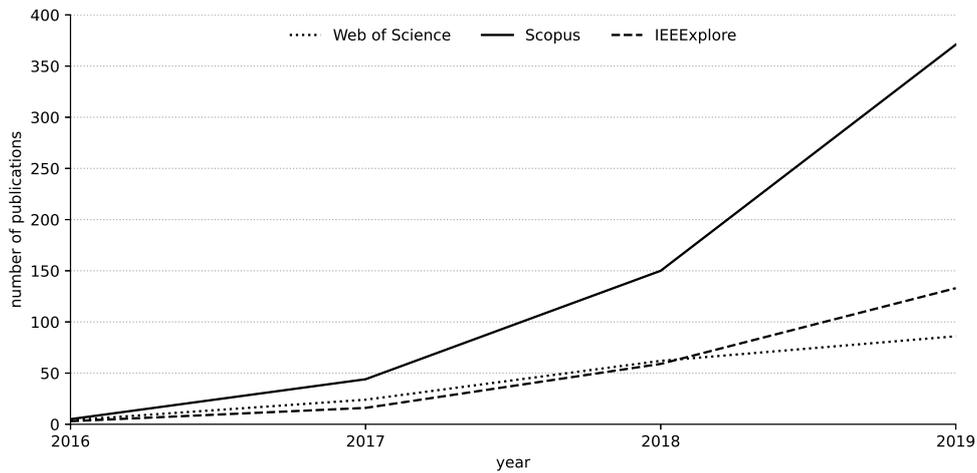}
    \caption{Evolution of the number of publications \emph{per year} retrieved from the keyword "\emph{fake news detection}" according to \emph{Web of Science}, \emph{Scopus} and \emph{IEEExplore}.}
    \label{fig:wos}
\end{figure}

The raising interest in the fake news detection domain might easily be noticed by checking how many scientific articles have concerned this topic, according to commonly-used and relevant databases. Such metrics are shown in Figure \ref{fig:wos} that depicts the number of publications on fake news detection per year and database. According to that, in the \emph{Scopus} database there are 5 articles associated to the '\emph{fake news detection}' keyword and published in 2016, 44 in 2017 , 150 in 2018 and 371 in 2019. In the \emph{Web of Science} database there are 4 articles published in 2016, 24 in 2017, 62 in 2018 and finally 86 in 2019. There is a similar when looking at the \emph{IEEExplore} database, stating that there were 3, 16, 59 and 133 articles published respectively in 2016, 2017, 2018 and 2019.

In addition to the published papers, another key metric to check the interest of the research community and funding agencies on a certain topic is the number of funded projects in competitive calls. According to this idea, a list of the research EU-funded projects can be seen in Table \ref{tab:EUprojects}. Data have been compiled from CORDIS\footnote{\url{https://cordis.europa.eu/projects/enl}} database (July 2020) searching the terms: '\emph{fake news}', '\emph{disinformation}', and '\emph{deepfake}'.

\begin{table}[ht]
\caption{EU-funded research projects. Data extracted from \textsc{cordis} database in July 2020.}\vspace{.5em}
    \centering
        \begin{tabular}{lcl}
        \toprule
        \bfseries \textsc{year} & \bfseries \textsc{number} & \bfseries \textsc{project acronyms}\\
        \midrule
        \oldstylenums{2014} & 1 & \textsc{pheme}\\
        \oldstylenums{2015} & 0 & ---\\
        \oldstylenums{2016} & 5 & \textsc{comprop}, \textsc{debunker}, \textsc{dante}, \textsc{encase}, \emph{InVID}\\
        \oldstylenums{2017} & 1 & \textsc{botfind}\\
        \oldstylenums{2018} & 7 & \textsc{dynnet}, \textsc{fandango}, \emph{GoodNews}, \textsc{jolt}, \emph{SocialTruth}, \textsc{soma}, \emph{WeVerify}\\
        \oldstylenums{2019} & 8 & \emph{Factmata}, \textsc{fakeology}, \textsc{digiact}, \emph{Media and Conspiracy}, \textsc{newtral}, \textsc{qi}, \textsc{rusinform}, \textsc{truthcheck}\\
        \oldstylenums{2020} & 5 & \textsc{d}i\textsc{ced}, \emph{mistrust}, \emph{News in groups}, \textsc{printout}, \textsc{radicalisation}\\
        \bottomrule
        \end{tabular}
\label{tab:EUprojects}
\end{table}

Among other EU projects, the \emph{Social Truth} project can be highlighted. It is a project funded by the \emph{Horizon 2020} R\&D program and it addresses the burning issue of fake news. Its purpose is to deal with this matter in a way that would let vendors to lock-in the solution, build up the trust and reputation by means of the \emph{blockchain} technology, incorporate \emph{Lifelong Learning Machines} which are capable of spotting the paradigm changes of false information and provide a handy digital companion which would be able to support the individuals in their verifying the services they use, from within web browsers they utilise. To meet this objective, ICT engineers, data scientists and \emph{blockchain} experts belonging to both the industry and the academia, together with end-users and use-cases providers have created a consortium in order to combine their efforts. Further details may be found in~\cite{choras2019socialtruth}.

It can be concluded that the European Union (EU) works hard in order to combat online misinformation and to educate the society. 
As a result, increasing amounts of economic resources are being invested.

\subsection{Image tampering datasets}
There are multiple datasets of modified images available online. One of them is CASIA dataset (in fact, two version of this dataset: CASIA ITDE \emph{v1.0} and CASIA ITDE \emph{v1.0}) described in \cite{dong2013casia} and \cite{zheng2019puf}. The ground truth input pictures are sourced from the CASIA ITDE image tampering detection evaluation (ITDE) \emph{v1.0} database; it comprises of images belonging to eight categories (animal, architecture, article, character, nature, plant, scene and texture), sized 384x256 or 256x384. Comparing datasets CASIA ITDE \emph{v1.0} and CASIA ITDE \emph{v2.0}, the newer one proves to be more challenging and comprehensive. It utilises post-processing, such as blurring or filtering of the tampered parts to render the manipulated images seem realistic to one's eye. In the CASIA ITDE \emph{v2.0} dataset, there can be a number of tampered versions for each genuine image. 

In accordance with CASIA ITDE \emph{v2.0}, the manipulated images are created by applying crop-and-paste operation in \emph{Adobe Photoshop} on the genuine pictures, and the altered areas might be of irregular shapes and various measurements, rotations or distortions. 

Among datasets of modified images available online, the one proposed in \cite{christlein2012evaluation} should also be enumerated. It contains unmodified/original images, unmodified/original images with JPEG compression, 1-to-1 splices (i.e. direct copy of snippet into image), splices with added Gaussian noise, splices with added compression artefacts, rotated copies, scaled copies, combined effects and copies that were pasted multiple times. Mostly, the subsets exist in downscaled versions as well. There are two image formats available: JPEG and TIFF, even though the TIFF format can have a size up to 30 GB.

The other dataset was provided by \emph{CVIP Group} working at \emph{Department of Industrial and Digital Innovation} (\emph{DIID}) of University of Palermo \cite{ardizzone2015copy}. The Dataset comprises medium-sized images (most of them 1000x700 or 700x1000) and is further divided into multiple datasets (\emph{D0, D1, D2}). The first dataset \emph{D0} is composed of 50 not compressed images with simply translated copies. For remaining two sets of images (\emph{D1, D2}), 20 not compressed images were selected, showing simple scenes (single object, simple background). \emph{D1} subset has been made by copy-pasting rotated elements, while \emph{D2} - scaled ones.

Next dataset was proposed in \cite{castro2020dataset}; it is the \emph{CG-1050} dataset which comprises 100 original images, 1050 tampered images and their corresponding masks. The dataset is divided into four directories: original, tampered and mask images, along with a description file. The directory of original images contains 15 colour and 85 grayscale images. The directory of tampered images comprises 1050 images obtained by one of the following methods of tampering: copy-move, cut-paste, retouching and colorizing.

There are also some datasets consisting on videos. As an example, the \emph{Deepfake Detection Challenge Dataset} can be listed~\cite{DFDC2020}. This dataset contains 124k videos that have been modified using eight facial modification algorithms. The dataset is useful for deepfake modification of videos.


\subsection{Fake news datasets}
 The LIAR dataset, where there are almost thirteen thousand manually labelled brief statements in varied context taken from the website \emph{polifact.com}, was introduced in~\cite{wang2017liar}. It contains the data collected over a span of a decade and marked as: \emph{pants-on-fire}, \emph{false}, \emph{barely-true}, \emph{half-true}, \emph{mostly-true}, and \emph{true}. The label distribution is fairly well-balanced: besides 1,050 \emph{pants-on-fire} cases, there are between 2,063 to 2,638 examples for each label.

The dataset used in~\cite{horne2017just} in fact consisted of three datasets: (\emph{i}) \emph{Buzzfeed} -- data collected from \emph{Facebook} concerning the US Presidential Election, (\emph{ii}) \emph{Political news} -- news taken from trusted sources (\emph{The Guardian}, \emph{BBC}, etc.), fake sources (\emph{Ending the Fed}, \emph{Inforwars}, etc.) and satire (\emph{The Onion}, \emph{SatireWire}, etc.) and (\emph{iii}) \emph{Burfoot and Baldwin} -- the dataset proposed in 2009, containing mostly real news stories. Unfortunately, the whole dataset is not well-balanced. It contains 4,111 real news, 110 fake news and 308 satire news.

The CREDBANK dataset was introduced in~\cite{mitra2015credbank}. It is an extensive, crowd-sourced dataset containing about 60 million tweets covering 96 days, beginning from October 2015. The tweets relate to more than 1,000 news events, with each of them checked for credibility by 30 editors from \emph{Amazon Mechanical Turk}.

The \emph{FakeNewsNet} repository, which is updated in a periodical manner, was proposed by the authors of~\cite{shu2018fakenewsnet}. This dataset contains the combination of news items (source, body, multimedia) and social background details (user profile, followers/followee) concerning fake and truthful materials, gathered from \emph{Snopes} and \emph{BuzzFeed}, that have been reposted and shared on \emph{Twitter}.

The next dataset is \emph{ISOT Fake News Dataset} that was described in~\cite{ahmed2018detecting}. It is very well-balanced: contains over 12,600 false and true news items each. The dataset was gathered using real world outlets; the truthful items were collected by crawling the articles from \emph{Reuters.com}. Conversely, the articles containing false information were picked up from diverse sources. The false news articles were gathered from unreliable web pages flagged by \emph{Politifact} (a US-based fact-checking entity) and \emph{Wikipedia}.

Another method for detecting fake news, called the multimodal one, was shown in~\cite{wang2018eann}. There, authors defined possible features that may be used during the analysis. They are: textual features (statistical or semantic), visual features (image analysis) and social context features (followers, hashtags, retweets). In the presented approach, textual and visual features were used for detecting fake news.

\begin{table}[!ht]
    \centering
    \caption{The review of existing datasets.}\vspace{.5em}
    \begin{tabular}{lrlr}
        \toprule
        \bfseries \textsc{dataset} & \multicolumn{2}{c}{ \bfseries \textsc{elements}} & \bfseries \textsc{citation}\\
        & \textsc{quantity}
        & \textsc{category}\\
        \midrule
        
        LIAR &
        \begin{tabular}{@{}r@{}}
        1,050 \\ 2,063 -- 2,638 
        \end{tabular}
        &
        \begin{tabular}{@{}l@{}}
        \emph{pants-on-fire} \\ \emph{others}
        \end{tabular}
        &
        \cite{wang2017liar} \\
        
        \midrule
        
        \begin{tabular}{@{}l@{}}
            Buzzfeed dataset \\ 
            + Political news dataset \\ 
            + Burfoot and Baldwin dataset 
        \end{tabular} &
         \begin{tabular}{@{}r@{}}
        4,111 \\ 110 \\ 308 
        \end{tabular}
        &
        \begin{tabular}{@{}l@{}}
        \emph{real news} \\ \emph{fake} \\ \emph{satire}
        \end{tabular}
        &

        \cite{horne2017just} \\
        
        \midrule
        
        CREDBANK &
        \multicolumn{2}{c}{\emph{60 million tweets}} &
        \cite{mitra2015credbank} \\
        
        \midrule
        
        \emph{FakeNewsNet} &
        \multicolumn{2}{c}{\emph{no data}} &
        \cite{shu2018fakenewsnet} \\
        
        \midrule
        
        \emph{ISOT Fake News Dataset} &
        
         \begin{tabular}{@{}r@{}}
        > 12,600 \\ > 12,600 
        \end{tabular}
        &
        \begin{tabular}{@{}l@{}}
        \emph{real news} \\ \emph{fake}
        \end{tabular}
        &
        \cite{ahmed2018detecting} \\
        
        \midrule
        
        \emph{Twitter + Weibo} &
         \begin{tabular}{@{}r@{}}
        12,647 \\ 10,805 \\ 10,042 
        \end{tabular}
        &
        \begin{tabular}{@{}l@{}}
        \emph{real news} \\ \emph{fake} \\ \emph{images}
        \end{tabular}
        &
        \cite{wang2018eann} \\\bottomrule

    \end{tabular}
    \label{tab:my_label}
\end{table}




\section{Conclusions, further challenges and way forward}
This section draws the main conclusions of the present research, regarding the application of advanced ML techniques. Additionally, open challenges in the disinformation arena are pointed out.

\subsection{Streaming nature of fake news}

It should be underlined that most of the papers addressing fake news detection ignore the streaming nature of this task. The profile of items labelled as fake news might shift over time because the spreaders of false news are conscious of the fact that automatic detection systems could detect them. As a result, they try to avoid their messages being identified as fake news by changing some of their characteristics. Therefore, in order to continuously detect them, ML-driven systems have to react to these changes, known as \emph{concept drift} \cite{Krawczyk:2017}. It requires to equip the detection systems with mechanisms able to adapt to changes. Only a few papers have attempted to develop fake news detection algorithms taking into consideration the streaming nature of the data \cite{Ksieniewicz:2019}. Though a number of researchers noted social media should be considered as data streams, only Wang and Terano \cite{Wang:2015} used appropriate techniques for data stream analysis. Nevertheless, their method is restricted to quite short streams and probably did not reflect the non-stationary nature of the data. Ksieniewicz et al. \cite{Ksieniewicz:2020} employed NLP techniques and treated incoming messages as a non-stationary data stream. The computer experiments on real-life false news datasets prove the usefulness of the suggested approach.

\subsection{Lifelong learning solutions}

Lifelong ML systems may transcend the restrictions of canonical learning algorithms that require a substantial set of training samples and are fit for isolated single-task learning \cite{Chen:2018}. Key features which should be developed in the systems of this kind in order to take advantage of prior learned knowledge comprise feature modeling, saving what had been learnt from past tasks, transferring the knowledge to upcoming learning tasks, updating the previously learnt things and user feedback. 
Additionally, the idea of a '\emph{task}' which is present in several conventional definitions \cite{Pentina:2015} of lifelong ML models, proves difficult to specify in numerous real-life setups (oftentimes, it seems hard to tell when a given task ends and the subsequent one begins).
One of the major troubles is the dilemma of \emph{stability and plasticity}, i.e. the situation where the learning systems must compromise between learning new information without forgetting the previous one \cite{Jin:2004}. It is visible in the catastrophic forgetting phenomenon, which is described as a neural network forgetting the previously learned information entirely, after having been exposed to new information. 

We believe that lifelong learning systems and methods would perfectly fit the fake news problem where content, style, language and types of fake news change rapidly. 

\subsection{Explainability of ML-based fake news detection systems}
Additional point which must be considered at present time is the explainability of ML and ML-based fake news detection methods and systems. Unfortunately, numerous scientists and systems architects utilise deep-learning capacities (along with other black-box ML techniques) in performing detecting or prediction assignments. However, the outcome produced by the algorithms is given with no explanation. Explainability concerns the extent to which a human is able to comprehend and explain (in a literal way) the internal mechanics driving the AI/ML systems.

Indeed, for the ML-based fake detection methods to be successful and widely trusted by different communities (journalism, security etc.), the relevant decision-makers in a realistic environment need the answer to the following question: what is the reason for the system to give certain answers \cite{choras2020machine}?

\subsection{The emergence of deepfakes}
It is worth mentioning that, going one step further in this subject, a new phenomenon has recently appeared, referred to as \emph{deepfakes}. Initially, they could be defined as hyper-realistic movie files applying face swaps which do not leave much trace of having been tampered with~\cite{deepfakeChawla}. This ultimate manipulation now consists in the generation of fake media resources by using AI face-swap technology. The contents of graphical deepfakes (both pictures and videos) are mainly the people whose faces are substituted. On the other hand, there are also deepfakes recordings in which the voices of people are simulated. Although there can be potential productive uses of deepfakes~\cite{deepfakeWesterlund}, they may also imply negative economic effects~\cite{deepfakeTourism}, as well as severe legal ones\footnote{\url{https://spectrum.ieee.org/tech-talk/computing/software/what-are-deepfakes-how-are-they-created}}.

Although an audit has revealed that the software to generate deepfake videos is still hard to use\footnote{\url{https://spectrum.ieee.org/tech-talk/computing/software/the-worlds-first-audit-of-deepfake-videos-and-tools-on-the-open-web}}, there is an increase in such fake contents, not only affecting celebrities~\footnote{\url{https://www.bbc.com/news/av/technology-40598465}}, but also less-known people\footnote{\url{https://www.bbc.co.uk/bbcthree/article/779c940c-c6c3-4d6b-9104-bef9459cc8bd}}$^,$\footnote{\url{https://www.theguardian.com/technology/2020/jan/13/what-are-deepfakes-and-how-can-you-spot-them}}. As some authors have previously stated, technology will play a keystone role in fighting deepfakes ~\cite{deepfakeMaras}. In this sense, authors in \cite{ciftci2020hearts} have very recently presented an approach to accurately detect fake portrait videos ($97.29\%$ accuracy) as well as to find out the particular generative model underlying a deep fake based on spatiotemporal patterns present in biological signals, under the assumption that a synthetic person, for instance, does not show a similar pattern of heart beat in comparison to the real one. Nevertheless, contributions are required from other fields such as legal, educational and political ones~\cite{deepfakeWesterlund,deepfakePitt}.

As for some other open challenges related to cybersecurity, fake news and deepfakes require increasing the resources spent on detection technology; identification rates must be increased while the sophistication of disinformation continuously grows~\cite{deepfakeTourism}.


\subsection{Final remarks}
This work presents the results obtained from a comprehensive and systematic study of research papers, projects and initiatives concerning detecting fake news (online disinformation). Our goal was to show current and possible trends in this needed area of research in the computer science field due to the demands of societies from countries worldwide. Additionally, available resources (methods, datasets, etc.) to research in this topic have been thoroughly analysed. 

In addition to the analysed previous work, the present study is aimed at motivating researchers to take up challenges in this domain, that increasingly impact current societies. More precisely, challenges still to be addressed are identified in order to propose exploring them. 

\section*{Acknowledgement}
\noindent This work is supported by the SocialTruth project\footnote{\url{http://socialtruth.eu}}, which has received funding from the European Union’s Horizon 2020 research and innovation programme under grant agreement No. 825477.

\bibliographystyle{elsarticle-num-names}
\bibliography{bib}

\end{document}